\pdfoutput=1
\documentclass[11pt]{article}

\usepackage[]{acl}

\usepackage{times}
\usepackage{latexsym}

\usepackage[T1]{fontenc}
\usepackage[utf8]{inputenc}

\usepackage{microtype}

\usepackage{inconsolata}

\usepackage{graphicx}

\usepackage{makecell}
\usepackage{booktabs}
\usepackage{multirow}
\usepackage{algorithm}
\usepackage{algorithmic}
\usepackage{enumitem}
\usepackage{xcolor}
\usepackage{subfigure}
\usepackage{diagbox} 
\usepackage{amsmath}
\usepackage{amsfonts}
\usepackage{bbding}

\newif\ifshowcomments
\showcommentstrue   % 将该行注释掉来隐藏注释

\ifshowcomments
    \newcommand{\jc}[1]{{\color{purple} [jiachen: #1]}}
    
    \newcommand{\zcm}[1]{{\color{red}[congming: #1]}}
\else
    \newcommand{\jc}[1]{}
    \newcommand{\zcm}[1]{}
\fi

\title{Retrieval-Augmented Process Reward Model for Generalizable Mathematical Reasoning}

\author{
  Jiachen Zhu\textsuperscript{\rm 1}, Congmin Zheng\textsuperscript{\rm 1}, Jianghao Lin\textsuperscript{\rm 1}, Kounianhua Du\textsuperscript{\rm 1}\\\textbf{Ying Wen}\textsuperscript{\rm 1}$^{\ast}$, \textbf{Yong Yu}\textsuperscript{\rm 1}, \textbf{Jun Wang}\textsuperscript{\rm 2}, \textbf{Weinan Zhang}\textsuperscript{\rm 1}\thanks{*Corresponding author}\\
  \textsuperscript{\rm 1}Shanghai Jiao Tong University, \textsuperscript{\rm 2}University College London \\ 
    \texttt{\{gebro13,desp.zcm,chiangel,kounianhuadu,ying.wen,wnzhang\}@sjtu.edu.cn, }\\
    \texttt{yyu@apex.sjtu.edu.cn}\\
    \texttt{jun.wang@cs.ucl.ac.uk}
    }
\begin{document}
\maketitle
\begin{abstract}
While large language models (LLMs) have significantly advanced mathematical reasoning, Process Reward Models (PRMs) have been developed to evaluate the logical validity of reasoning steps. However, PRMs still struggle with out-of-distribution (OOD) challenges. This paper identifies key OOD issues, including step OOD—caused by differences in reasoning patterns across model types and sizes—and question OOD, which arises from dataset shifts between training data and real-world problems. To address these issues, we introduce Retrieval-Augmented Process Reward Model (RetrievalPRM), a novel framework designed to tackle these OOD issues. By utilizing a two-stage retrieval-enhanced mechanism, RetrievalPRM retrieves semantically similar questions and steps as a warmup, enhancing PRM’s ability to evaluate target steps and improving generalization and reasoning consistency across different models and problem types. Our extensive experiments demonstrate that RetrievalPRM outperforms existing baselines across multiple real-world datasets.  Our open-source contributions include a retrieval-enhanced dataset, a tuning framework for PRM training, and the RetrievalPRM model, establishing a new standard for PRM performance.
\end{abstract}

\section{Introduction}

% In recent years, Large Language Models (LLMs) have made remarkable advances in mathematical reasoning ~\cite{openai2023gpt,dubey2024llama,zhu2024deepseek,shao2024deepseekmath,yang2024qwen2}, yet they can make mistakes, such as miscalculations or logical errors, leading to wrong conclusions. Moreover, even when achieving correct final answers, these powerful models can still regularly make up plausible reasoning steps, where the final answers build upon flawed calculations or derivations, which undermine the reliability and trustworthiness of LLMs’ reasoning processes. To address these challenges, Process Reward Models (PRMs)~\cite{lightman2023let,wang2024math}, have emerged as a critical component in the realm of automated reasoning. PRMs extend beyond traditional outcome-based evaluation by explicitly assessing the coherence and logical validity of intermediate reasoning steps~\cite{gsm8k}. This paradigm aligns with human pedagogical practices, where educators emphasize not only final answers but also the quality of problem-solving strategies.

While large language models (LLMs) have advanced mathematical reasoning~\cite{openai2023gpt,dubey2024llama,zhu2024deepseek,shao2024deepseekmath,yang2024qwen2}, they remain prone to critical flaws: explicit errors (e.g., miscalculations, logical inconsistencies) and implicit risks where correct answers mask flawed intermediate steps. Even when final results are accurate, LLMs often generate plausible-but-incorrect reasoning chains, eroding trust in their problem-solving processes \cite{lightman2023let}. To address this, Process Reward Models (PRMs) \cite{lightman2023let,wang2024math} have been developed to rigorously evaluate the logical validity of intermediate steps \cite{gsm8k}, mirroring human pedagogical practices that prioritize reasoning quality over answer correctness.

\begin{figure}[t]
  \centering
  \vspace{-10pt}
  \includegraphics[width=0.48\textwidth]{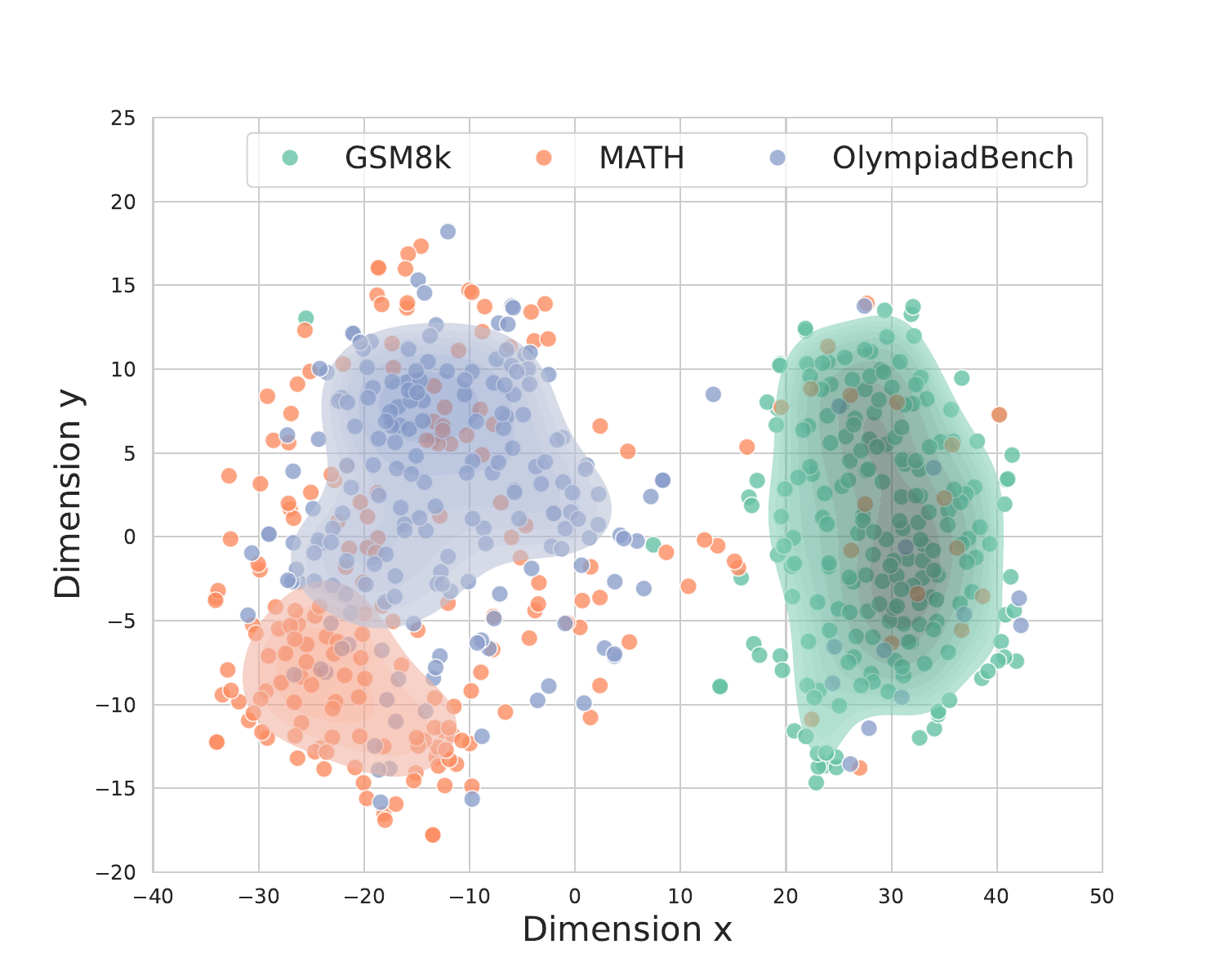}
  \vspace{-20pt}
  \caption{The distribution differences across three datasets: GSM8K, MATH and Olympiad. We use sentence-bert to encode these questions and perform t-sne visualization.}
  \label{fig: question OOD}
  \vspace{-20pt}
\end{figure}

Existing works~\cite{wang2024openropensourceframework, skyworkopeno12024,zheng2024processbench} frame PRM as a binary classification problem. They train PRM on open-source base LLMs such as Qwen~\cite{yang2024qwen2} or Llama~\cite{dubey2024llama} using human-annotated dataset ~\cite{lightman2023let} or automated process supervision method ~\cite{wang2024math,luo2024improvemathematicalreasoninglanguage,qin2024o1replicationjourneystrategic}. Although these approaches show great performance and empirical success, they still face kinds of out-of-distribution challenges. 
We believe the out-of-distribution (OOD) problem can be viewed from the following perspectives:  

\begin{figure*}[t]
  \centering
  \vspace{-30pt}
  \includegraphics[width=\textwidth]{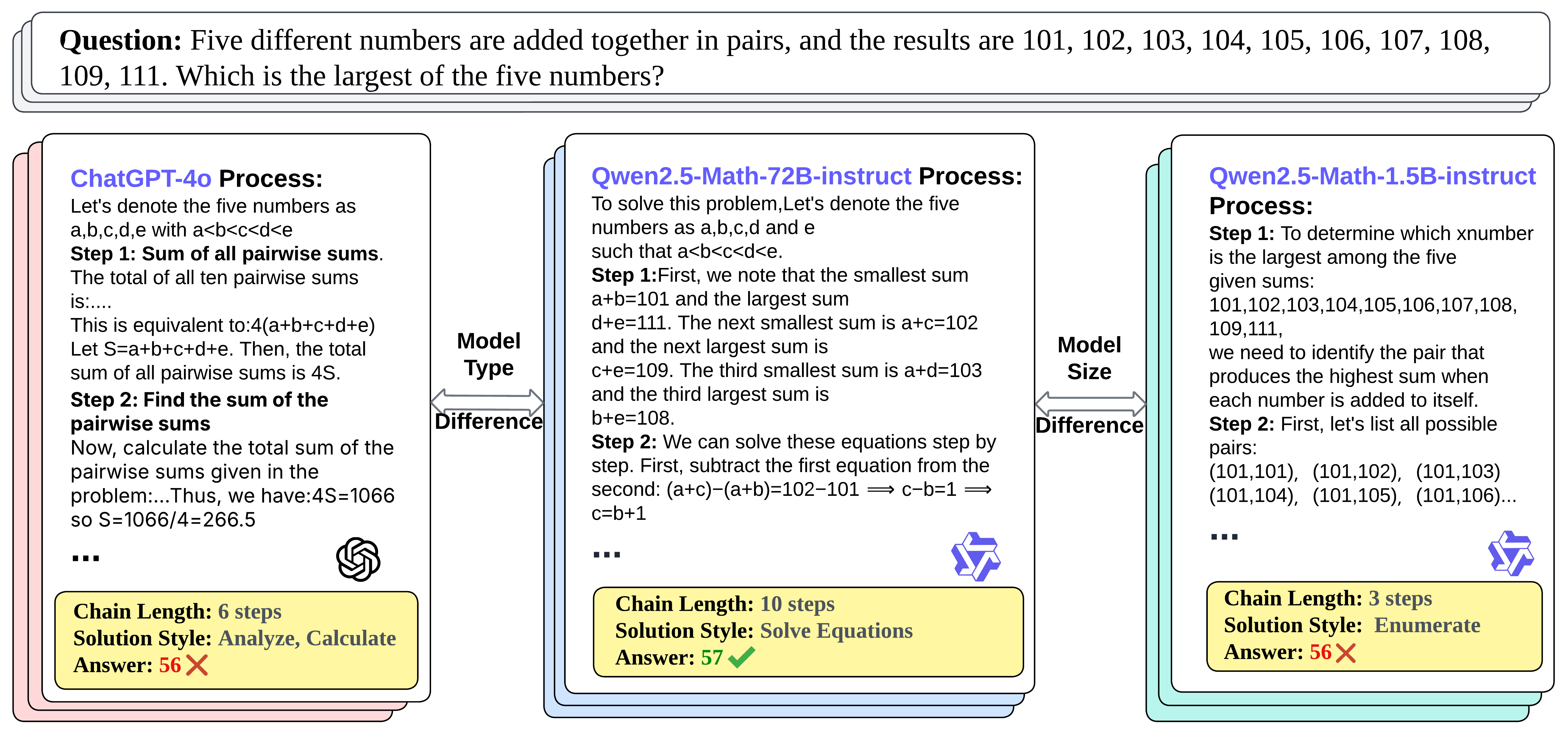}
  \vspace{-20pt}
  \caption{Processes and problem-solving ideas for the same question vary from different models with the perspectives of model types and model sizes. GPT tends to analyze and calculate, while Qwen-72B tends to solve equations. Qwen-1.5B is small and relatively weak. It can only enumerate, and its thinking chain is short, so its answers are also very wrong.}
  \label{fig: model OOD}
  \vspace{-10pt}
\end{figure*}

Firstly, \textbf{Step OOD} may occur because of different processes generated by different models. Due to the high cost of manual annotation, there are very few accurately labeled PRM expert datasets, such as PRM800K and ProcessBench, with processes generated by GPT~\cite{openai2023gpt} and Qwen~\cite{yang2024qwen2}, respectively. However, different model types (e.g., GPT, Qwen, Llama\cite{dubey2024llama}) approach problem-solving differently. As is shown in Figure~\ref{fig: model OOD}, when facing the same question, GPT-4o tends to analyze and calculate, while Qwen-72B tends to solve questions directly. They have different solution styles. Therefore, using process data generated by one model to train a PRM and then applying it to guide another model leads to an OOD issue. Moreover, models of different sizes also exhibit different reasoning processes. Larger models, like exceptional students, tend to have clearer and more accurate reasoning steps, while smaller models tend to have very short reasoning chains, as shown in Figure~\ref{fig: model OOD}.

Secondly, \textbf{Question OOD} emerges because of dataset shift. Current PRM datasets contain only a limited number of problems. For example, Math Shepherd and PRM800K cover problems from the GSM8K and MATH datasets, with GSM8K being at the elementary to middle school level and MATH at the high school to university level. However, real-world problems are far more diverse, such as those in the Olympic math competition dataset~\cite{he2024olympiadbenchchallengingbenchmarkpromoting}, leading to OOD issues in other datasets. As shown in the Figure~\ref{fig: question OOD}, we used Sentence-BERT~\cite{reimers2019sentence} to encode all the problems from the three datasets and visualized the distribution with t-SNE. It is evident that the distributions differ, and since both Olympic and MATH problems are typically from high school-level exams, they are semantically closer to each other than to GSM8K.

To address this issue, we propose a new framework, Retrieval Augmented Process Reward Model (\textbf{RetrievalPRM}), which leverages a Two-stage Retrieval-enhanced Mechanism to help PRMs solve the OOD problem. we retrieve relevant questions and steps in these two stages to address the issues of question OOD and step OOD, respectively. Specifically, when predicting a step for a given question, we select semantically similar questions based on their embeddings, placing them at the beginning of the entire prompt. Additionally, we select more fine-grained, similar steps and use them as references when predicting the correctness of the step. These retrieved questions and steps serve as a kind of warm-up for PRM, acting as example problems for reference. They not only help stimulate PRM’s potential by warming up but also allow the system to handle more difficult problems by identifying similarities, thus alleviating OOD issues. 
% Furthermore, we discuss the sample construction methods, putting multi-labels in one sample for efficiency and designing a Reference Step Attention Mask, where, when predicting the $n_{th}$ step's correctness, only the current node's references and the previous steps are visible, preventing the previous steps’ references from influencing the prediction.

\noindent Our main contributions are summarized as follows: 
\begin{itemize}[leftmargin=10pt]
    \item To the best of our knowledge, we are the first to highlight the key OOD problems in Process Reward Models (PRMs), particularly the question OOD and step OOD, which arise due to differences in reasoning patterns across model types (e.g., GPT, Qwen), model sizes (1.5B, 72B) and varying problem difficulties in real-world datasets.
    \item  We introduce the Retrieval-Augmented Process Reward Model (\textbf{RetrievalPRM}) framework, which utilizes a Two-stage Retrieval-enhanced Mechanism to address OOD issues by incorporating both Question-level Retrieval and Step-level Retrieval, thereby enhancing PRM's ability to generalize across diverse problem-solving scenarios.
    % \item We discuss the ways of sample construction methods and propose a novel Reference Step Attention Mask to prevent the influence of previous steps' references when predicting the $n_{th}$ step.
    \item We build a Retrieval-enhanced dataset for training PRM using RetrievalPRM framework. We have made our code publicly available.\footnote{https://anonymous.4open.science/r/RetrievalPRM-1C77} Our dataset\footnote{https://huggingface.co/datasets/gebro13/RetrievalPRM\_\\Dataset} and model\footnote{https://huggingface.co/gebro13/RetrievalPRM} are open-sourced.
    \item Extensive experiments on the ProcessBench~\cite{zheng2024processbench} on four public real-world datasets demonstrate that RetrievalPRM outperforms strong baselines and that the Out-of-distribution issue has been alleviated due to our retrieval approach. 
\end{itemize}
\section{Preliminary}
\label{sec:preliminary}

In this section, we formulate the whole problem and introduce PRM as a binary classification model.

\subsection{Problem Formulation}
We denote the Math dataset as $\mathcal{D}=\{(q_i,\mathbf{s}_i,\mathbf{y}_i)\}_{i=1}^N$, where $N$ is the number of data instances. 
The input $q_i$ is the $i^{th}$ Math question. $\mathbf{s}_i = \{ s^1_i,s^2_i,\ldots, s^{n_i}_i\}$ are the solution steps, where $n_i$ is the step number of solution $\mathbf{s}_i$. $\mathbf{y}_i = \{ y^1_i, y^2_i, \ldots, y^{n_i}_i \}$ and the label $y^j_i$ indicates the correctness from the $1^{st}$ step to the $j^{th}$ step.
\begin{equation}
	y^j_{i} = 
		\begin{cases}
		    1,~ (s^1_i,\ldots,s^j_i)~\text{is correct for} ~q_i; \\
			0,~ \text{otherwise.} 
		\end{cases}
\end{equation}
\subsection{ORM vs. PRM}
Outcome-supervised Reward Models are introduced (ORM) by~\cite{gsm8k}, where verifiers are trained for judging the final correctness of generated solutions. ORM only predicts the final label $\hat{y}^{n_i}_i$, which can be formulated as 
\begin{equation}
	\forall{i},\hat{y}^{n_i}_{i} = \text{ORM}(q_i,s^1_i,\ldots,s^{n_i}_i).
\end{equation}

Building on this, the concept of process reward models (PRM) is introduced as a more granular
 and transparent approach. Not only does PRM evaluate the final solutions but it also assesses intermediate processes, where $\hat{y}^{j}_{i}$ represents the predicted label for the $j^{th}$ step by PRM.
\begin{equation}
	\forall{i,j}, \hat{y}^{j}_{i} = \text{PRM} (q_i,s^1_i,\ldots,s^{j}_i).
\end{equation}

\subsection{Large Language Model for PRM scoring}
\label{llm as for prm scoing}
 
When directly adopting LLMs as the PRM for scoring, we need to convert the data $(q_i,\mathbf{s}_i,\mathbf{y}_i)$ with a hard prompt template. The whole template example is illustrated in Appendix~\ref{app:prompts}.
% \begin{figure}
%   \centering
%   % \vspace{-10pt}
%   \includegraphics[width=0.5\textwidth]{imgs/Prompt Illusttration.pdf}
%   \vspace{-10pt}
%   \caption{The illustration of PRM input template.}
%   \vspace{-10pt}
%   \label{fig:prompt template example}
% \end{figure}

The textual input consists of the question $q_i$ and steps $\mathbf{s}_i$, followed by a binary question about the correctness of these steps. 

To obtain the floating-point correctness estimation $\hat{y}_i^{j}\in[0,1]$ instead of discrete word tokens '+' or '-', we apply bidimensional softmax over the corresponding logits of the binary key answer tokens (ie., + \& -) from LLMs to accomplish the  correctness estimation during evaluation:
\begin{equation}
    \hat{y}_i^j=\frac{\exp(l_{i,\text{+}})}{\exp(l_{i,\text{+}})+\exp(l_{i,\text{-}})}\in(0,1).
\end{equation}
where $l_{i,\text{+}}$ and $l_{i,\text{-}}$ are the logits of token + and - in the  $i^{th}$ instance, respectively. 

It is important to note that the estimated PRM scoring $\hat{y}_i^j$ is used solely for evaluation on the testing set. If training is involved, we maintain the standard instruction tuning and causal language modeling paradigm for LLMs. In this way, we don't need to replace the language model head with binary classification head which is the last layer of LLM.

\section{Methodology}

In this section, we introduce our proposed \textbf{\textit{RetrievalPRM}} framework in detail.

\begin{figure*}[t]
  \centering
  \vspace{-30pt}
  \includegraphics[width=1.0\textwidth]{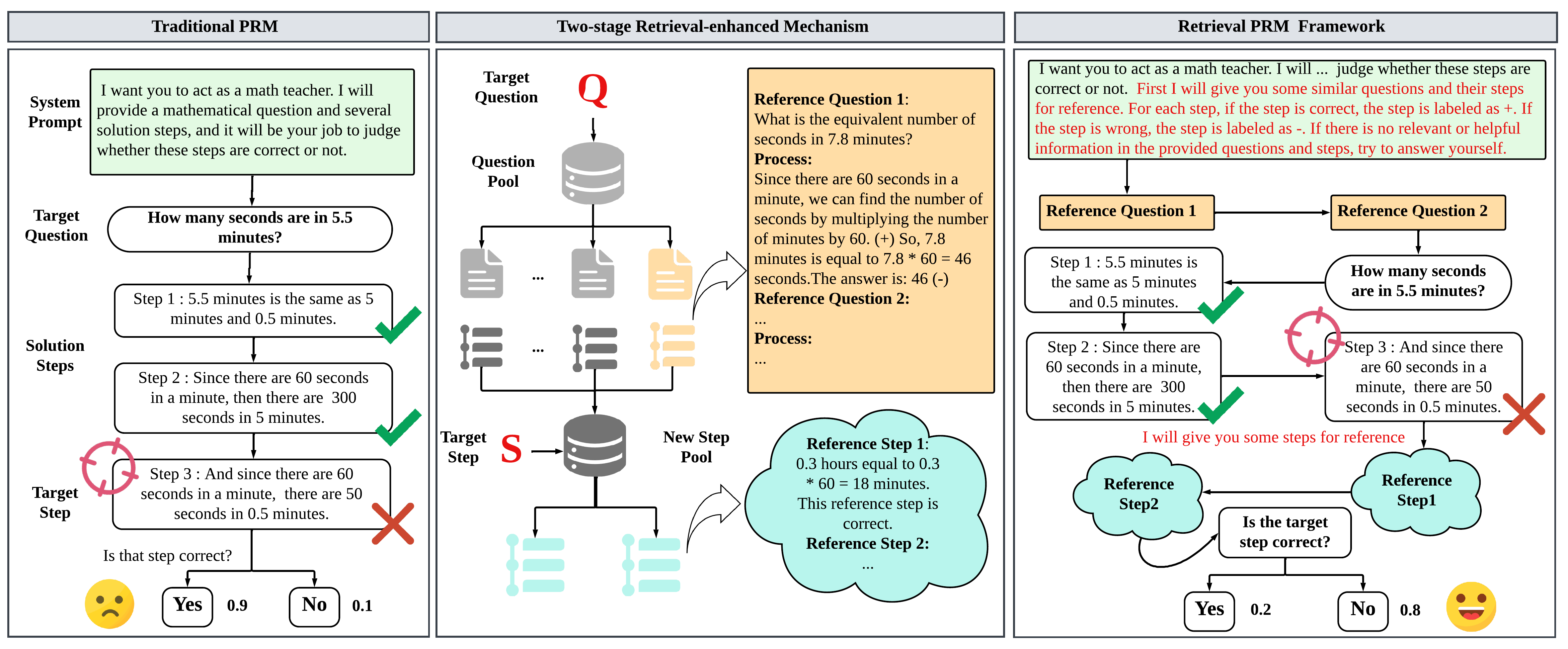}
  \vspace{-20pt}
  \caption{The model structure of our proposed RetrievalPRM framework and its difference with traditional PRM. We design a Two-stage Retrieval Module to retrieve reference questions and steps in each stage. }
  \label{fig: Framework}
  \vspace{-10pt}
\end{figure*}

\subsection{Overview of RetrievalPRM}

The RetrievalPRM is developed to address the problem of out-of-distribution (OOD) scenarios in mathematical problem-solving, specifically focusing on both question OOD and step OOD. According to Figure~\ref{fig: Framework}, traditional PRM models are constrained by predefined solution steps and are unable to handle unseen questions or steps effectively, especially when the problem context shifts or the solution process deviates from previously seen examples. RetrievalPRM overcomes this challenge by incorporating a Two-stage Retrieval-enhanced Mechanism that dynamically fetches relevant questions and steps from a large pool of questions and their solutions. These retrieved questions and steps serve as a kind of warm-up for PRM, acting as example problems for reference. They not only help stimulate PRM's potential by warming up but also allow the system to handle more difficult problems by identifying similarities.

% By integrating retrieval mechanisms, the model can access a wide variety of solutions and apply them to both familiar and novel questions, effectively dealing with the distributional shift in problem types (question OOD) and solution steps (step OOD). This makes RetrievalPRM particularly effective in real-world scenarios where problem contexts are diverse and rapidly evolving.

\subsection{Two-stage Retrieval-enhanced Mechanism}
The core of RetrievalPRM is the Two-stage Retrieval-enhanced Mechanism, which consists of two key phases: Question-level Retrieval and Step-level Retrieval.

\subsubsection{Question-level Retrieval}
\label{question-level retrieval}
The first stage of retrieval tackles the question OOD issue. As is shown in Figure~\ref{fig: Framework}, the retrieval pool is the question database $\mathbb{D}_q = \{q_i\}_{i=1}^N$. During retrieval process, we treat:
\begin{itemize}[leftmargin=10pt]
    \item Query: the target question $q_t$.
    \item Key: all $q_i$ in the retrieval pool.
    \item Value: all the $(q_i,\mathbf{s}_i)$ pair in the retrieval pool.
\end{itemize}
We calculate their similarities $<q_i,q_t>$ to match the most similar n questions. Specifically, all questions will first pass through a Sentence-BERT model to encode questions and obtain their semantic representations.
\begin{equation}
    \{e_{q_i}\}_{i=1}^N = \text{SentenceBERT}(\{q_i\}_{i=1}^N)
\end{equation}
where $e_{q_i} \in \mathbb{R}^{D}$ is the embedding vector of the question $q_i$.

And then all the embeddings undergo Principle Component Analysis (PCA)~\cite{kurita2021principal} for dimensionality reduction to extract the most important dimensions. 
\begin{equation}
    \{e'_{q_i}\}_{i=1}^N = \text{PCA}( \{e_{q_i}\}_{i=1}^N )
\end{equation}
where $e'_{q_i} \in \mathbb{R}^d$ is the embedding after dimension reduction.

Finally, we compute the cosine similarity between the target question and the entire question pool, selecting the top-\textit{k} most similar questions and inputting them into the text.
\begin{equation}
    \begin{aligned}
    \langle q_i,q_t\rangle &= \frac{e'_{q_t} \cdot  e'_{q_i}}{|e'_{q_t}|\cdot|e'_{q_i}|}.\\
    \end{aligned}
\end{equation}

Now we sort the vector $\{\langle q_i,q_t\rangle\}_{i=1}^N$ of similarity and choose top-\textit{k} $(q_i,\mathbf{s}_i)$ pairs as reference questions $q_r$ and put them in RetrievalPRM's input together with the target question. Furthermore, we store all the solutions $\{\mathbf{s}_i\}_{i=1}^{m}$ of top-\textit{m} ($m>k$) questions in a new database to conduct a further step-level retrieval. 

\subsubsection{Step-level Retrieval}

We place step-level retrieval in the second stage of the two-stage retrieval process, rather than as a separate module, for two key reasons:

Firstly, for a solution to be meaningful, both the question and the steps must be similar. For example, two different types of questions might both use the letter "p" to represent an unknown variable, but in some problems, "p" represents a prime number, while in others, it represents probability. This results in steps that may appear similar but have entirely different meanings, rendering the retrieved steps potentially unhelpful.

Secondly, since there are many possible solutions to a question, this leads to a large number of steps. If the majority of these steps are irrelevant, the time spent calculating similarities becomes inefficient. By placing step-level retrieval in the second stage, we can save both time and computational resources.

Therefore, after retrieving the top-\textit{m} most similar questions, we inject all their solutions into a new steps database $\mathbb{D}_s$. Then, we use the target step as the query to retrieve reference steps from this new database. The similarity for retrieval is still calculated using Sentence-BERT, PCA, and cosine similarity, as mentioned in ~\ref{question-level retrieval}.

\subsection{Retrieval-based System Prompt}
In RetrievalPRM,  The system prompt serves as the instruction set for the model, framing the problem and directing it to evaluate each step of the solution.
Besides the traditional system prompt for PRM, the Retrieval-based System Prompt (RetSP) is extended with additional instructions, as shown in the red sentence in Figure~\ref{fig: Framework}, which encourages the model to leverage knowledge from reference questions. For example, we inform PRM that step labels "+" and "-" represent correct and incorrect steps, respectively. At the same time, to avoid noise, we specify that if the reference question or step contains no relevant or helpful information, it should not be considered. These retrieval-based system prompts give PRM a more flexible thinking process, enabling it to actively decide whether to use retrieval-based knowledge.

% For example, the prompt might ask the model to judge if the steps in the target solution are correct or not, based on the reference steps. If the model determines that a step is incorrect, the system prompts it to adjust or correct the solution based on the retrieved references. This retrieval-driven approach not only improves the model’s performance in judgment tasks but also allows the model to generalize better by learning from a diverse set of reference solutions, rather than being limited to pre-programmed rules or static examples.

% This methodology outlines the essential components and workflow of the RetrievalPRM model. The use of a retrieval-based system enhances its ability to solve mathematical problems by dynamically sourcing relevant references to assess solution correctness.
We define reference questions of $q_i$ as $\mathbf{q}_i^{r}$ and reference steps as $\mathbf{s}_i^r$. The whole input $\mathbf{x}_i^j$ of predicting the $j_{th}$ step of $q_i$ in RetrievalPRM can be formulated as:
\begin{equation}
\label{input template}
    \begin{aligned}
    \mathbf{x}^j_i = (RetSP,\mathbf{q}^r_i,&q_i,
    s^1_i,\ldots,s^{j-1}_i,\mathbf{s}^r_i,s^j_i,y^j_i),\\
    \hat{y}^j_i = & \text{PRM}(\mathbf{x}^j_i)
    \end{aligned}
\end{equation}
where $s^j_i$ is the $j_{th}$ step of solution $\mathbf{s}_i$.

According to the input template above, it is worth noting that when predicting step n, we assume that steps 1 through n-1 are correct ~\cite{luo2024improvemathematicalreasoninglanguage,zheng2024processbench}. At this point, the most important task for PRM is to predict step n, so PRM can only access the reference steps for step n and cannot see the reference steps for steps $1\sim n-1$.

\section{Experiments}
\begin{table*}[h]
\centering    
\vspace{-10pt}
\caption{The performance of different models on ProcessBench. 
The best result is given in bold, and the second-best value is underlined. See Table~\ref{tab:All performance addition} in Appendix~\ref{app: supplementary results} for breakdown of evaluation results.}
% The symbol $\ast$ indicates a statistically significant improvement of RetrievalPRM over the best baseline with $p$-value < 0.01.}
\vspace{-8pt}

\label{tab:All performance}
\resizebox{1.0\textwidth}{!}{
\renewcommand\arraystretch{1.1}
\begin{tabular}{clccccccccc}
\hline
\multicolumn{2}{c}{\multirow{2}{*}{Model}} & \multicolumn{2}{c}{GSM8k} & \multicolumn{2}{c}{MATH} & \multicolumn{2}{c}{OlympiadBench} & \multicolumn{2}{c}{OmniMATH} & \multirow{2}{*}{Avg.F1}  \\ 
\cmidrule(r){3-4} \cmidrule(r){5-6} \cmidrule(r){7-8} \cmidrule(r){9-10}
\multicolumn{2}{c}{} & ArithACC & F1 & ArithACC & F1 & ArithACC & F1 & ArithACC & F1 &  \multicolumn{1}{c}{} \\ 
\hline 

\multicolumn{1}{c|}{\multirow{7}{*}{\makecell{Open-source \\ PRM}}}
& RetrievalPRM-7B(Ours) & \textbf{76.0} & \textbf{74.6} & \textbf{70.6}& \textbf{71.1} & \textbf{59.1} & \textbf{60.2} & \textbf{55.2} & \textbf{57.33} & \textbf{65.8} \\
\multicolumn{1}{c|}{\multirow{4}{*}{}} & Qwen2.5-Math-7B-PRM800K & \underline{73.5} & 68.2 &\underline{65.1} & \underline{62.6} & \underline{53.2} & \underline{50.7} & \underline{43.4} & \underline{44.3} &  \underline{56.5} \\
\multicolumn{1}{c|}{\multirow{4}{*}{}} & Skywork-PRM-7B & 71.6 & \underline{70.8} & 54.5 & 53.6 & 25.6 & 22.9 & 23.7 & 21.0 & 42.1 \\
\multicolumn{1}{c|}{\multirow{4}{*}{}} & Skywork-PRM-1.5B & 59.9 & 59.0 & 49.1 & 48.0 & 20.5 & 19.3 & 19.7 & 19.2 & 36.4 \\
\multicolumn{1}{c|}{\multirow{4}{*}{}} & Math-Shepherd-PRM-7B & 58.3 & 47.9 & 45.1 & 29.5 & 39.7 & 24.8 & 34.8 & 23.8 & 31.5 \\
\multicolumn{1}{c|}{\multirow{4}{*}{}} & RLHFlow-PRM-Mistral-8B & 62.3 & 50.4 & 42.1 & 33.4 & 22.3 & 13.8 & 19.1 & 15.8 & 28.4 \\
\multicolumn{1}{c|}{\multirow{4}{*}{}} & RLHFlow-PRM-Deepseek-8B & 56.9 & 38.8 & 45.1 & 33.8 & 26.5 & 16.9 & 23.2 & 16.9 & 26.6 \\

\hline

\multicolumn{1}{c|}{\multirow{17}{*}{\makecell{Language \\ Models \\ as Critic}}}
& QwQ-32B-Preview & \textbf{87.9} & \textbf{88.0} & \textbf{78.5} & \textbf{78.7} & \underline{59.2} & \textbf{57.8} &\textbf{61.1} & \textbf{61.3} & \textbf{71.5} \\
\multicolumn{1}{c|}{\multirow{4}{*}{}} & GPT-4o& 80.2 & 79.2& 63.4 &\underline{63.6} & 50.1&51.4& 50.1 & \underline{53.5} & \underline{61.9} \\
\multicolumn{1}{c|}{\multirow{4}{*}{}} & Qwen2.5-72B-Instruct & 77.9 & 76.2 & \underline{65.4} & 61.8 & \textbf{59.8} & \underline{54.6} & 55.1 & 52.2 &  61.2 \\

\multicolumn{1}{c|}{\multirow{4}{*}{}} & Llama-3.3-70B-Instruct & \underline{83.7} & \underline{82.9} & 63.7 & 59.4 & 54.3 & 46.7 & 51.0 & 43.0 & 58.0 \\
\multicolumn{1}{c|}{\multirow{4}{*}{}} & Qwen2.5-Coder-32B-Instruct & 72.0 & 68.9 & 64.5 & 60.1 & 57.0 & 48.9 & 52.5 & 46.3 & 56.1 \\
\multicolumn{1}{c|}{\multirow{4}{*}{}} & Llama-3.1-70B-Instruct & 75.3 & 74.9 & 52.6 & 48.2 & 50.0 & 46.7 & 43.2 & 41.0 & 52.7 \\
\multicolumn{1}{c|}{\multirow{4}{*}{}} & Qwen2.5-14B-Instruct & 72.3 & 69.3 & 59.2 & 53.3 & 50.2 & 45.0 & 43.5 & 41.3 & 52.2 \\
\multicolumn{1}{c|}{\multirow{4}{*}{}} & Qwen2-72B-Instruct & 67.8 & 67.6 & 52.3 & 49.2 & 43.3 & 42.1 & 39.3 & 40.2 & 49.8 \\
\multicolumn{1}{c|}{\multirow{4}{*}{}} & Qwen2.5-32B-Instruct & 70.6 & 65.6 & 61.9 & 53.1 & 53.5 & 40.0 & 47.7 & 38.3 & 49.3 \\
\multicolumn{1}{c|}{\multirow{4}{*}{}} & Qwen2.5-Math-72B-Instruct & 70.3 & 65.8 & 59.6 & 52.1 & 56.1 & 32.5 & 55.1 & 31.7 & 45.5 \\
\multicolumn{1}{c|}{\multirow{4}{*}{}} & Qwen2.5-Coder-14B-Instruct & 61.9 & 50.1 & 54.2 & 39.9 & 51.4 & 34.0 & \underline{55.6} & 27.3 & 37.8 \\
\multicolumn{1}{c|}{\multirow{4}{*}{}} & Qwen2.5-7B-Instruct & 37.8 & 36.5 & 36.9 & 36.6 & 29.9 & 29.7 & 27.3 & 27.4 & 32.6 \\
\multicolumn{1}{c|}{\multirow{4}{*}{}} & Meta-Llama-3-70B-Instruct & 62.4 & 52.2 & 48.3 & 22.8 & 46.2 & 21.2 & 44.8 & 20.0 & 29.1 \\
\multicolumn{1}{c|}{\multirow{4}{*}{}} & Qwen2.5-Math-7B-Instruct & 54.4 & 26.8 & 50.3 & 25.7 & 43.1 & 14.2 & 41.6 & 12.7 & 19.9 \\
\multicolumn{1}{c|}{\multirow{4}{*}{}} & Qwen2-7B-Instruct & 25.1 & 8.4 & 20.4 & 19.0 & 16.1 & 14.7 & 13.8 & 12.1 & 13.6 \\
\multicolumn{1}{c|}{\multirow{4}{*}{}} & Meta-Llama-3-8B-Instruct & 27.1 & 13.1 & 17.3 & 13.8 & 14.2 & 4.8 & 19.7 & 12.6 & 11.1 \\
\multicolumn{1}{c|}{\multirow{4}{*}{}} & Qwen2.5-Coder-7B-Instruct & 49.1 & 14.3 & 46.3 & 6.5 & 47.2 & 4.1 & 48.9 & 1.8 & 6.7 \\
\multicolumn{1}{c|}{\multirow{4}{*}{}} & Llama-3.1-8B-Instruct & 27.3 & 10.9 & 20.5 & 5.1 & 16.0 & 2.8 & 15.0 & 1.6 & 5.1 \\

\hline
\end{tabular}
\vspace{-5pt}
}
\end{table*}

\begin{table*}[h]
\centering    

\vspace{-5pt}
\caption{The performance of different variants of RetrievalPRM on ProcessBench.  We remove different components of RetrievalPRM to evaluate the contribution of each part to the model. The best result is given in bold, and the second-best value is underlined. See Table~\ref{tab:ablation performance addition} in Appendix~\ref{app: supplementary results} for breakdown of evaluation results.
}
\vspace{-8pt}

\label{tab:ablation performance}
\resizebox{1.0\textwidth}{!}{
\renewcommand\arraystretch{1.1}
\begin{tabular}{cccccccccccc}
% \toprule
\hline

\multicolumn{2}{c}{Retrieval Components} & \multicolumn{2}{c}{GSM8k} &\multicolumn{2}{c}{MATH} &\multicolumn{2}{c}{OlympiadBench}&\multicolumn{2}{c}{OmniMATH}&\multirow{2}{*}{Avg.F1}\\ 

 \cmidrule(r){3-4} \cmidrule(r){5-6}  \cmidrule(r){7-8}  \cmidrule(r){9-10}
Question-level &Step-level & ArithACC & F1 & ArithACC & F1 & ArithACC & F1 & ArithACC & F1 &  \multicolumn{1}{c}{}  \\ 
   \hline

\checkmark &\checkmark& \underline{76.0}&\underline{74.6} &\underline{70.6}& \underline{71.1 }& \textbf{59.1}&\textbf{ 60.2}&\textbf{55.2}& \textbf{57.3}&\textbf{65.8}\\
\checkmark&$\times$ &\textbf{77.8} &\textbf{74.9} &\textbf{70.7}&\textbf{71.2}&\underline{58.4}&\underline{59.8} &50.5&54.4&\underline{65.0}\\
$\times$& \checkmark&73.8 &67.5 &69.5 & 69.2& 58.2& 58.9&\underline{52.2} & \underline{56.3} &63.0\\
$\times$&$\times$& 71.0&65.6 & 67.3 & 67.5&54.3 & 55.8 & 47.2&50.9 &59.9\\

\hline   
\end{tabular}
\vspace{-5pt}
}

\end{table*}

\begin{figure*}[t]
  \centering
  \vspace{-30pt}
  \includegraphics[width=1.0\textwidth]{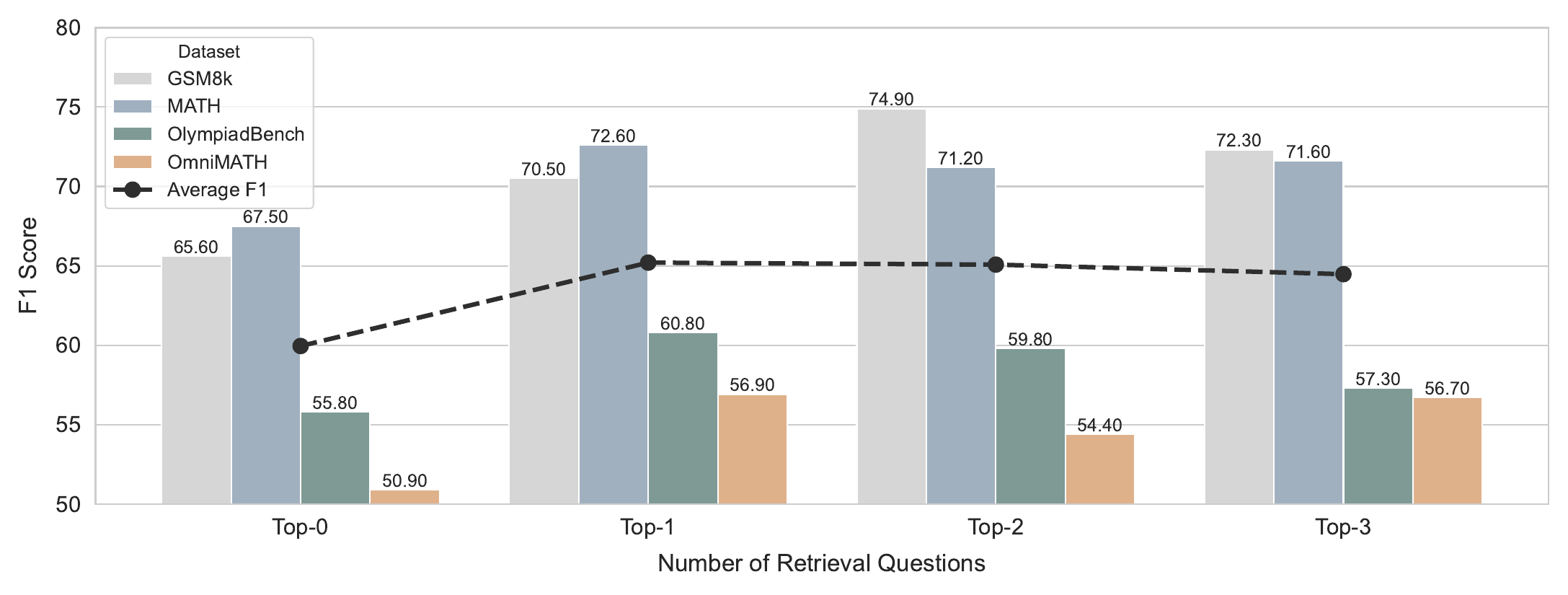}
  \vspace{-25pt}
  \caption{We show the F1 scores of Retrieval-PRM on four datasets and their average, as the number of retrieval questions varies. Specifically, Top-0 means no retrieval questions. }
  \label{fig: Hyperparameter}
  \vspace{-10pt}
\end{figure*}

In this section, we present the experimental settings and results. Our implementation code of RetrievalPRM is publicly available.
% The following discussions are guided by three research questions.

% \noindent\textbf{RQ1} Does TR-PRM outperform existing baselines?

% \noindent\textbf{RQ2} What is the impact of different components in TR-PRM?

% \noindent\textbf{RQ3} What impact does the number of retrieval questions have on the performance of the model?
\subsection{Experiment Setup}
\subsubsection{Datasets}
Datasets are categorized into two kinds: Math reasoning datasets, and prm training datasets.

\noindent \textbf{Math Reasoning Datasets}

We conduct experiments on four public and widely used datasets in mathematical reasoning tasks: \textit{GSM8K}~\cite{gsm8k} which contains math problems from elementary to middle school, \textit{MATH}~\cite{hendrycks2021measuring} which contains math problems from basic to university level, \textit{OlympiadBench}~\cite{he2024olympiadbenchchallengingbenchmarkpromoting} which involves questions from the Mathematical Olympiad, \textit{Omni-MATH}~\cite{gao2024omnimathuniversalolympiadlevel} which covers multi-domain high-difficulty problems. Further details are provided in Appendix~\ref{app:datasets}.

Except for GSM8K, which focuses on grade school math problems, the other three datasets feature problems of competition or Olympiad-level difficulty.

\noindent \textbf{PRM training datasets}

We conduct experiments on two publicly available datasets for PRM:

\textit{PRM800K}~\cite{lightman2023let}: Based on the MATH dataset, it contains 800,000 manually annotated step-level correctness labels for training the Process Reward Model. It relies on expensive manual annotations.

\textit{Math-Shepherd}~\cite{wang2024math}: It generates 400,000 machine-annotated step-level labels (covering MATH and GSM8K datasets) by automatically building process supervision data, without manual annotation.

\subsubsection{Evaluation Metrics}
We evaluate our model in a public PRM benchmark ProcessBench~\cite{zheng2024processbench}. The aim is to judge whether PRM can find the first wrong step. It divides data into two parts: samples with incorrect and correct final answers and then conducts harmonic mean on the accuracy of these two parts to get the final F1-score. Moreover, we think since the sample number of each part isn't balanced, We add an additional metric: weighted arithmetic mean of these two parts, which is shown in Table~\ref{tab:All performance} as ArithACC.

\subsubsection{Baselines}
Following~\cite{zheng2024processbench}, we divide all baselines into two parts:

(1) \textit{Open-source PRM}, including Skywork~\cite{skyworkopeno12024}, Qwen2.5-PRM~\cite{zheng2024processbench}, Math-Shepherd~\cite{wang2024math} and RLHFlow~\cite{xiong2024rlhflowmath}. These models are binary classification PRMs.

(2) \textit{Language Models as Critic}, including Llama~\cite{dubey2024llama}, Qwen2~\cite{yang2024qwen2}, Qwen2.5~\cite{qwen2.5}, Qwen2.5-MATH~\cite{yang2024qwen25mathtechnicalreportmathematical}, Qwen2.5-Coder~\cite{hui2024qwen25codertechnicalreport}, GPT-4o~\cite{openai2024gpt4ocard}. These models are promoted to judge the steps with the help of majority voting.

Further details of these baselines are provided in Appendix~\ref{app:baselines} due to article length limitations. 

\subsubsection{Implementation Details}
Details like base models, hyperparameters, prompts, and training sizes are provided in Appendix~\ref{app:implementation details} due to the article length limitations.

\subsection{Overall Performance}
We evaluate RetrievalPRM against existing baselines on ProcessBench, and the results are presented in Table~\ref{tab:All performance}. The findings are as follows:

\begin{itemize}[leftmargin=10pt]
    \item RetrievalPRM-7B surpasses all open-source PRM baselines, achieving the highest performance. Notably, the most significant improvement is observed on OmniMATH, the most challenging dataset, with performance gains increasing as dataset difficulty rises. This phenomenon may stem from the fact that most baseline PRMs are trained on human- or machine-annotated datasets such as PRM800K or Math-Shepherd, which primarily focus on GSM8K or MATH and exhibit OOD issues when applied to more complex datasets. In contrast, our RetrievalPRM effectively mitigates the OOD problem through its retrieval-based approach, demonstrating the efficacy of our Two-stage Retrieval-enhanced Mechanism.
    \item When comparing models of different scales, RetrievalPRM outperforms all evaluated language models, including Qwen2.5-72B-Instruct and Llama3.3-70B-Instruct, with the sole exception of QwQ-32B-Preview. Remarkably, RetrievalPRM achieves this with a model size of just 7B. This highlights that PRMs, being both lightweight and task-specific, maintain strong competitiveness and potential compared to LLMs as critics.
\end{itemize}
\subsection{Ablation Study}
We analyze two main components in the Two-stage Retrieval-enhanced Mechanism: \emph{Question-level Retrieval} and \emph{Step-level Retrieval}—through the following ablations:

\noindent\textbf{RetrievalPRM (Ours)}: The complete version of our proposed method.

\noindent\textbf{RetrievalPRM (w/o Step-level Retrieval)}: This variant retains only the Question-level Retrieval, removing Step-level Retrieval during both training and inference.

\noindent\textbf{RetrievalPRM (w/o Question-level Retrieval)}: This variant retains only the Step-level Retrieval, removing Question-level Retrieval during both training and inference.

\noindent\textbf{RetrievalPRM (w/o Question-level and Step-level Retrieval)}: In this variant, both Question-level and Step-level Retrieval are removed during training and inference.

The performance of these variants is presented in Table~\ref{tab:ablation performance}, from which we can draw the following observations:
\begin{itemize}[leftmargin=10pt]
    \item The performance of RetrievalPRM (w/o Step-level Retrieval) remains almost identical to that of RetrievalPRM on GSM8K and MATH but exhibits a slight decline on OlympiadBench and OmniMATH. This can be attributed to the fact that Step-level Retrieval information is partially absorbed by Question-level Retrieval. As a result, Question-level Retrieval alone may be sufficiently effective for relatively easy datasets, as the reference steps it provides contain adequate knowledge for step prediction. However, for more challenging datasets, Step-level Retrieval becomes significantly more crucial, as it offers finer-grained guidance essential for handling complex problem-solving processes.
    \item RetrievalPRM (w/o Question-level Retrieval) shows lower performance, as it relies solely on Step-level Retrieval. The model lacks knowledge of reference questions, which is useful to alleviate question OOD, restricting its overall performance. 
    \item RetrievalPRM (w/o both Retrieval) performs the worst, which is expected, demonstrating the effectiveness of both question-level and Step-level Retrieval.
\end{itemize}

\subsection{Hyperparameter Study}

Figure~\ref{fig: Hyperparameter} illustrates the impact of the number of retrieval questions on the model's performance. The findings are as follows:

Compared to Top-0, where no retrieval questions are used, models that incorporate retrieval questions show improved performance, highlighting the importance of Question-level Retrieval. It inspires us that Reference questions are important for PRM to get warmup, no matter how many reference questions there are.

The performance of Top-3 exhibits a slight decline, potentially due to two factors: (1) An excessive number of reference questions may lead to an overly long input prompt, making it difficult for PRMs to comprehend or extract key information effectively. (2) A limited retrieval pool might result in later reference questions being less relevant than earlier ones, increasing the likelihood of misjudgments in the model’s predictions.

\section{Related Works}
\subsection{Process Reward Models}
% Process reward models (PRMs) have been shown to have advancement over traditional outcome reward models (ORMs)\cite{gsm8k} in training models’ process-level reasoning accuracy and improving their long-process reasoning abilities~\cite{lightman2023let,uesato2022solvingmathwordproblems}. More and more PRMs have been proposed for use in process-level RLHF~\cite{wang2024math,qin2024o1replicationjourneystrategic,xia2025evaluatingmathematicalreasoningaccuracy,skyworkopeno12024}. \cite{lightman2023let} released a large amount of labeled data at the human-annotated process level, providing great research opportunities for multi-step reasoning. \cite{wang2024math} introduces a self-supervised automatic data generation and PRM training pipeline that can automatically generate the process-level label. \cite{xia2025evaluatingmathematicalreasoningaccuracy} utilize PRM as an auto evaluator to assess the multistep reasoning accuracy of LMs. Due to the emergence of massive work on PRM training and data curating, many PRMs~\cite{skyworkopeno12024,xiong2024rlhflowmath,sun2024easytohardgeneralizationscalablealignment,gao2024llmcriticshelpcatch,wang2024openropensourceframework} have been proposed. In addition, some works focus on using natural language feedback from LLM as a reward~\cite{mcaleese2024llmcriticshelpcatch,zhang2024generativeverifiersrewardmodeling,gao2024llmcriticshelpcatch}, which are called critic models.
Process reward models (PRMs) have demonstrated significant advantages over traditional outcome reward models (ORMs)~\cite{gsm8k} in enhancing process-level reasoning accuracy and improving long-process reasoning abilities in model training~\cite{lightman2023let,uesato2022solvingmathwordproblems}. A growing number of PRMs have been proposed for application in process-level reinforcement learning with human feedback (RLHF)~\cite{wang2024math,qin2024o1replicationjourneystrategic,xia2025evaluatingmathematicalreasoningaccuracy,skyworkopeno12024}. For instance, \citet{lightman2023let} made a substantial contribution by releasing a large set of human-annotated data at the process level, opening up new research opportunities for multi-step reasoning. 

Additionally, \citet{wang2024math} introduces an automatic, self-supervised pipeline for generating process-level labels and training PRMs, enabling efficient data generation. \citet{xia2025evaluatingmathematicalreasoningaccuracy} employs PRMs as automatic evaluators to assess the accuracy of multi-step reasoning in language models (LMs). With the surge in PRM-focused research and data curation, numerous PRMs~\cite{skyworkopeno12024,xiong2024rlhflowmath,sun2024easytohardgeneralizationscalablealignment,gao2024llmcriticshelpcatch,wang2024openropensourceframework} have been proposed. Additionally, several studies focus on leveraging natural language feedback from large language models (LLMs) as rewards, which are termed critic models~\cite{mcaleese2024llmcriticshelpcatch,zhang2024generativeverifiersrewardmodeling,gao2024llmcriticshelpcatch}.

However, most existing PRMs trained on math datasets such as GSM8K and MATH inevitably encounter Out-of-distribution issues, which can be divided into two categories: \textbf{question OOD}, where PRMs trained on simpler or medium-difficulty datasets lack understanding of questions from more challenging datasets, and \textbf{step OOD}, where different base models and model sizes in LLMs lead to different step distributions for the same question. This is reflected in differences in chain length, problem-solving approaches, and methods. To address these issues, we propose the RetrievalPRM framework to tackle the OOD problems encountered in the current PRM field, achieving promising results.

\subsection{Retrieval-Augmented Generation}

Retrieval-augmented generation (RAG) enhances language models by dynamically integrating external knowledge, pioneered by~\cite{lewis2021retrievalaugmentedgenerationknowledgeintensivenlp} through their joint retrieval-generation architecture. Subsequent advances refined this paradigm. \citet{guu2020realmretrievalaugmentedlanguagemodel} introduced REALM to co-train retrieval and generation modules via masked language modeling, while~\citet{izacard2021leveragingpassageretrievalgenerative} proposed Fusion-in-Decoder (FiD) to process multi-document contexts efficiently. Research further optimized retrieval precision through dense passage embeddings~\cite{karpukhin2020densepassageretrievalopendomain} and scaled retrieval to web-level corpora~\cite{borgeaud2022improvinglanguagemodelsretrieving}. 
\section{Conclusion}
In this paper, we have addressed the significant out-of-distribution (OOD) challenges faced by Process Reward Models (PRMs), particularly step OOD and question OOD. By introducing the Retrieval Augmented Process Reward Model (RetrievalPRM), we propose an effective solution that leverages a Two-stage Retrieval-enhanced Mechanism to improve the generalization of PRMs across diverse models and problems. Extensive experiments on multiple real-world datasets have shown that RetrievalPRM consistently outperforms existing methods, highlighting its effectiveness in tackling OOD issues. 

\section{Limitation}
RetrievalPRM has two main limitations. Firstly, the retrieval pool is only constructed from PRM800K and Math-Shepherd at present, which is relatively small and limits the diversity and breadth of the mathematical problems. Second, using Sentence-BERT to embed questions and steps struggles to capture the full complexity of mathematical problems as semantic similarity doesn't mean knowledge similarity in Math problems. As a result, the naive cosine similarity calculated through embeddings may fail to accurately reflect the true similarity between two questions.

\bibliography{custom}

\appendix

\section{Baselines}
\label{app:baselines}
\subsection{Open-source PRM}
\begin{figure*}
  \centering
  % \vspace{-10pt}
  \includegraphics[width=\textwidth]{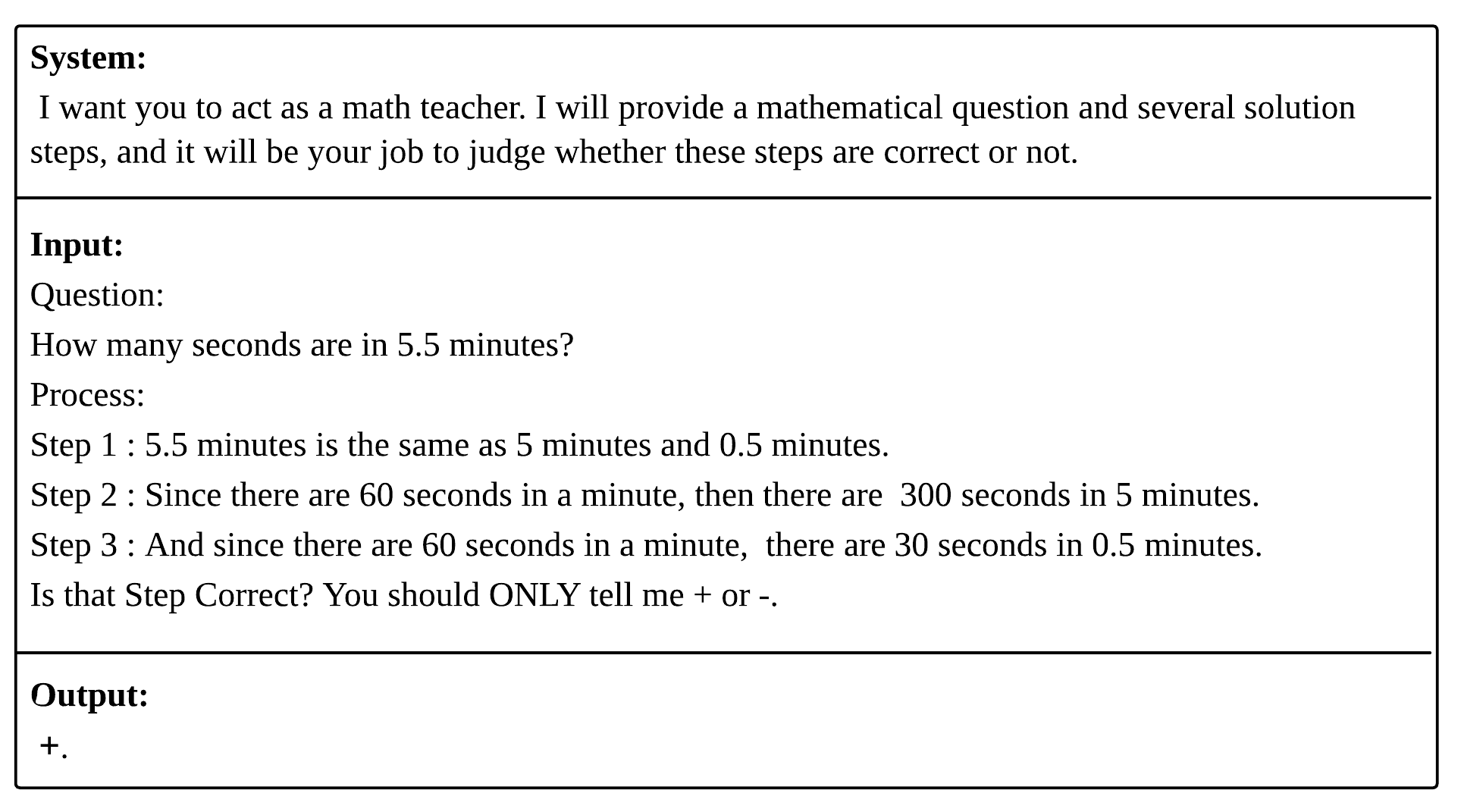}
  \vspace{-20pt}
  \caption{The illustration of PRM input template.}
  % \vspace{-10pt}
  \label{fig:prompt template example}
\end{figure*}
\begin{figure*}
  \centering
  % \vspace{-10pt}
  \includegraphics[width=\textwidth]{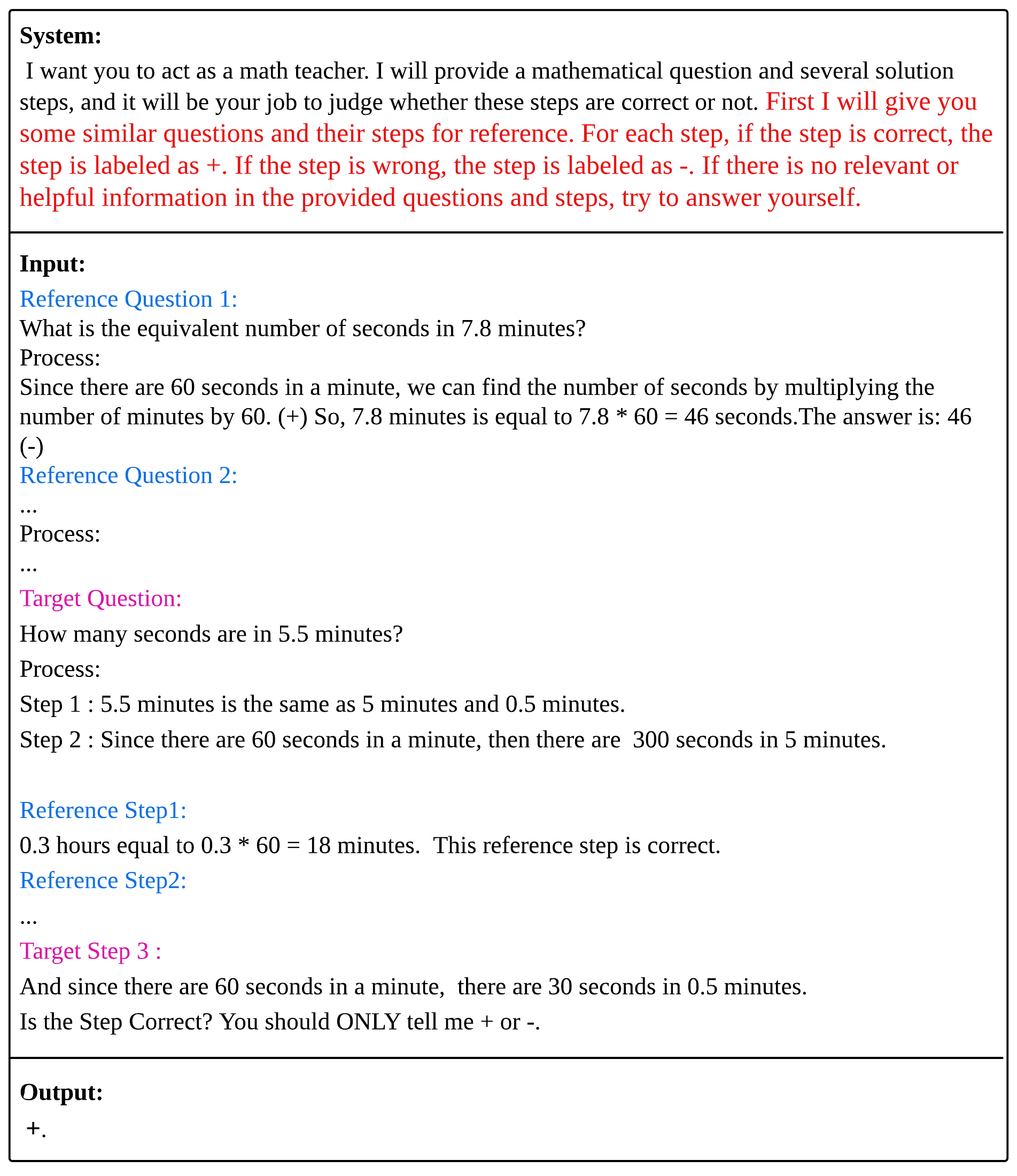}
  \vspace{-20pt}
  \caption{The illustration of RetrievalPRM input template.}
  % \vspace{-10pt}
  \label{fig:prompt template example2}
\end{figure*}

\begin{itemize}[leftmargin=10pt] 
\item Skywork-PRM~\cite{skyworkopeno12024} is a Qwen2.5-Math-based PRM published by KunLun. 
\item Qwen2.5-PRM~\cite{zheng2024processbench} is trained by fine-tuning the Qwen2.5-Math-7B-Instruct model on the PRM800K dataset. 
\item Math-Shepherd~\cite{wang2024math} generates process labels for each step by estimating the empirical probability that a given step leads to the correct final answer and trains a PRM based on their published dataset. 
\item RLHFlow-PRM~\cite{xiong2024rlhflowmath} is an 8-billion-parameter reward model trained with process supervision. 
\end{itemize}

\subsection{Language Models as Critic}
\begin{itemize}[leftmargin=10pt] 
\item Llama~\cite{dubey2024llama} is an open-source model developed by Meta (formerly Facebook), designed for natural language understanding and generation tasks. 
\item Qwen2~\cite{yang2024qwen2} is a large language model developed by Alibaba Cloud, offering multilingual support and strong capabilities in language understanding and generation. \item Qwen2.5~\cite{qwen2.5} is an advanced iteration of the Qwen series, pretrained on 18 trillion tokens, enhancing knowledge retention, programming, and mathematical reasoning. 
\item Qwen2.5-MATH~\cite{yang2024qwen25mathtechnicalreportmathematical} is a specialized model for mathematical problem-solving, trained on extensive math-focused data and incorporating Chain-of-Thought (CoT) and Tool-Integrated Reasoning (TIR). 
\item Qwen2.5-Coder~\cite{hui2024qwen25codertechnicalreport} is a programming-oriented model trained on 5.5 trillion code-related tokens, excelling in code generation, debugging, and multilingual programming tasks. 
\item GPT-4o~\cite{openai2024gpt4ocard} is a multimodal AI model developed by OpenAI that processes and generates text, audio, and images in real-time, with enhanced speed and natural interaction capabilities. 
\end{itemize}

\section{Implementation Details}
\label{app:implementation details}
\subsection{Basemodel and Training hyperparameters}

We selected Qwen-2.5-Math-7b-instruct\cite{qwen2.5} as the foundational large language model (LLM) for our experiments. All computations were performed using H100 GPUs. To enhance training resource efficiency, we employed Parameter-Efficient Fine-tuning techniques LoRA. The LoRA configuration was set with a rank of 32, an alpha value of 64, and dropout set to 0.1. LoRA update matrices were specifically applied to the query and value projection matrices within the attention blocks. 

We use PRM800K as our training data and both PRM800K and Math-Shepherd as our retrieval pool. The training process was carried out with batch sizes chosen from \(\{64, 128, 256, 512\}\) and initial learning rates selected from \(\{1\times 10^{-3},1\times 10^{-4}, 3 \times 10^{-4}, 1\times 10^{-5}, 3 \times 10^{-5}\}\) using a linear scheduler.

\subsection{Prompts}
\label{app:prompts}
In this section, we show our training prompts for PRM in details as is shown in Figure~\ref{fig:prompt template example} and Figure~\ref{fig:prompt template example2}.
% \begin{figure*}
%   \centering
%   % \vspace{-10pt}
%   \includegraphics[width=\textwidth]{imgs/Prompt Illusttration.pdf}
%   \vspace{-20pt}
%   \caption{The illustration of PRM input template.}
%   % \vspace{-10pt}
%   \label{fig:prompt template example}
% \end{figure*}

\section{Datasets}
\label{app:datasets}

\textit{GSM8K}~\cite{gsm8k}: Grade School Math is a dataset for basic to intermediate math problems, covering arithmetic, algebra, geometry and other fields. Its difficulty is suitable for math problems in elementary to middle school.

\textit{MATH}~\cite{hendrycks2021measuring}: The MATH dataset contains a variety of math problems from basic to university level, covering multiple mathematical fields such as algebra, geometry, calculus, number theory, etc.

\textit{OlympiadBench}~\cite{he2024olympiadbenchchallengingbenchmarkpromoting}: The Olympiadbench dataset contains questions from the Mathematical Olympiad. The questions are of high difficulty and involve complex combinatorial mathematics, number theory, geometry and other advanced mathematical fields.

\textit{Omni-MATH}~\cite{gao2024omnimathuniversalolympiadlevel}: Omni-MATH is a general Olympiad-level mathematics benchmark dataset for large language models, covering multi-domain and high-difficulty mathematics problems, and is designed to evaluate the reasoning ability of models in various mathematical fields.

Except for GSM8K, which focuses on grade school math problems, the other three datasets feature problems of competition or Olympiad-level difficulty.
\section{Supplementary Evaluation Results}
\label{app: supplementary results}
In this section, we show the breakdown of our main results in Table~\ref{tab:All performance addition} and ablation results in Table~\ref{tab:ablation performance addition}

\begin{table*}[h]
\centering    

\vspace{-10pt}
\caption{Breakdown of evaluation results of different models on ProcessBench. 
The best result is given in bold, and the second-best value is underlined. 
}
\vspace{-5pt}

\label{tab:All performance addition}
\resizebox{1.0\textwidth}{!}{
\renewcommand\arraystretch{1.1}
\begin{tabular}{cccccccccccccc}
% \toprule
\hline

\multicolumn{2}{c}{\multirow{2}{*}{Model}} & \multicolumn{3}{c}{GSM8k} &\multicolumn{3}{c}{MATH} &\multicolumn{3}{c}{OlympiadBench}&\multicolumn{3}{c}{OmniMATH}\\ 

 \cmidrule(r){3-5} \cmidrule(r){6-8}  \cmidrule(r){9-11}  \cmidrule(r){12-14}
\multicolumn{2}{c}{} & error  & correct & F1 & error  & correct & F1 & error  & correct & F1 & error  & correct & F1 \\ 
   \hline

\multicolumn{1}{c|}{\multirow{7}{*}{\makecell{Open-source \\ PRM}}}
& \multicolumn{1}{l}{RetrievalPRM-7B(Ours)} & 64.7 &88.1&\textbf{ 74.6} &67.2 & 75.6& \textbf{71.1 }& 56.0 & 65.2&\textbf{ 60.2}&52.8 & 62.65& \textbf{57.33}\\
\multicolumn{1}{c|}{\multirow{4}{*}{}} & \multicolumn{1}{l}{Qwen2.5-Math-7B-PRM800K} & 53.1 & 95.3 & 68.2 & 48.0 & 90.1 & \underline{62.6} & 35.7 & 87.3 & \underline{50.7} & 29.8 & 86.3 & \underline{44.3} \\
\multicolumn{1}{c|}{\multirow{4}{*}{}} & \multicolumn{1}{l}{Skywork-PRM-7B} & 61.8 & 82.9 & \underline{70.8} & 43.8 & 69.2 & 53.6 & 17.9 & 31.9 & 22.9 & 14.0 & 41.9 & 21.0\\
\multicolumn{1}{c|}{\multirow{4}{*}{}} &\multicolumn{1}{l}{ RLHFlow-PRM-Mistral-8B} & 33.8 & 99.0 & 50.4 & 21.7 & 72.2 & 33.4 & 8.2 & 43.1 & 13.8 & 9.6 & 45.2 & 15.8 \\
\multicolumn{1}{c|}{\multirow{4}{*}{}} &\multicolumn{1}{l}{ RLHFlow-PRM-Deepseek-8B} & 24.2 & 98.4 & 38.8 & 21.4 & 80.0 & 33.8 & 10.1 & 51.0 & 16.9 & 10.1 & 51.9 & 16.9 \\
\multicolumn{1}{c|}{\multirow{4}{*}{}} &\multicolumn{1}{l}{ Skywork-PRM-1.5B} & 50.2 & 71.5 & 59.0 & 37.9 & 65.3 & 48.0 & 15.4 & 26.0 & 19.3 & 13.6 & 32.8 & 19.2\\
\multicolumn{1}{c|}{\multirow{4}{*}{}} & \multicolumn{1}{l}{Math-Shepherd-PRM-7B} & 32.4 & 91.7 & 47.9 & 18.0 & 82.0 & 29.5 & 15.0 & 71.1 & 24.8 & 14.2 & 73.0 & 23.8 \\
\hline

\multicolumn{1}{c|}{\multirow{17}{*}{\makecell{Language \\ Models}}}
& \multicolumn{1}{l}{QwQ-32B-Preview} & 81.6 & 95.3 & \textbf{88.0} & 78.1 & 79.3 & \textbf{78.7} & 61.4 & 54.6 & \textbf{57.8} & 55.7 & 68.0 & \textbf{61.3}\\
\multicolumn{1}{c|}{\multirow{13}{*}{}} &\multicolumn{1}{l}{GPT-4o}& 70.0& 91.2 & 79.2 & 54.4 & 76.6 &\underline{63.6} & 45.8 & 58.4 & 51.4 & 45.2 &\underline{53.5}&\underline{61.9} \\ 
\multicolumn{1}{c|}{\multirow{13}{*}{}} &\multicolumn{1}{l}{ Qwen2.5-72B-Instruct }& 62.8 & 96.9 & 76.2 & 46.3 & 93.1 & 61.8 & 38.7 & 92.6 & \underline{54.6} & 36.6 & 90.9 & 52.2 \\ 
\multicolumn{1}{c|}{\multirow{13}{*}{}} &\multicolumn{1}{l}{ Llama-3.3-70B-Instruct }& 72.5 & 96.9 & \underline{82.9} & 43.3 & 94.6 & 59.4 & 31.0 & 94.1 & 46.7 & 28.2 & 90.5 & 43.0 \\
\multicolumn{1}{c|}{\multirow{13}{*}{}} & \multicolumn{1}{l}{Qwen2.5-32B-Instruct} & 49.3 & 97.9 & 65.6 & 36.7 & 95.8 & 53.1 & 25.3 & 95.9 & 40.0 & 24.1 & 92.5 & 38.3\\ 
\multicolumn{1}{c|}{\multirow{13}{*}{}} & \multicolumn{1}{l}{Qwen2.5-14B-Instruct} & 54.6 & 94.8 & 69.3 & 38.4 & 87.4 & 53.3 & 31.5 & 78.8 & 45.0 & 28.3 & 76.3 & 41.3 \\ 
\multicolumn{1}{c|}{\multirow{13}{*}{}} & \multicolumn{1}{l}{Qwen2.5-Coder-32B-Instruct} & 54.1 & 94.8 & 68.9 & 44.9 & 90.6 & 60.1 & 33.4 & 91.2 & 48.9 & 31.5 & 87.6 & 46.3 \\
\multicolumn{1}{c|}{\multirow{13}{*}{}} &\multicolumn{1}{l}{ Qwen2.5-Coder-14B-Instruct }& 33.8 & 96.4 & 50.1 & 25.4 & 92.4 & 39.9 & 20.7 & 94.1 & 34.0 & 15.9 & 94.2 & 27.3 \\
\multicolumn{1}{c|}{\multirow{13}{*}{}} & \multicolumn{1}{l}{Qwen2.5-Coder-7B-Instruct} & 7.7 & 100.0 & 14.3 & 3.4 & 98.3 & 6.5 & 2.1 & 99.1 & 4.1 & 0.9 & 98.3 & 1.8\\
\multicolumn{1}{c|}{\multirow{13}{*}{}} &\multicolumn{1}{l}{ Qwen2.5-Math-72B-Instruct} & 49.8 & 96.9 & 65.8 & 36.0 & 94.3 & 52.1 & 19.5 & 97.3 & 32.5 & 19.0 & 96.3 & 31.7 \\
\multicolumn{1}{c|}{\multirow{13}{*}{}} &\multicolumn{1}{l}{ Qwen2.5-Math-7B-Instruct} & 15.5 & 100.0 & 26.8 & 14.8 & 96.8 & 25.7 & 7.7 & 91.7 & 14.2 & 6.9 & 88.0 & 12.7 \\
\multicolumn{1}{c|}{\multirow{13}{*}{}} & \multicolumn{1}{l}{Llama-3.1-70B-Instruct} & 64.3 & 89.6 & 74.9 & 35.4 & 75.6 & 48.2 & 35.1 & 69.9 & 46.7 & 30.7 & 61.8 & 41.0 \\
\multicolumn{1}{c|}{\multirow{13}{*}{}} &\multicolumn{1}{l}{Meta-Llama-3-70B-Instruct} & 35.7 & 96.9 & 52.2 & 13.0 & 93.3 & 22.8 & 12.0 & 92.0 & 21.2 & 11.2 & 91.7 & 20.0 \\
\multicolumn{1}{c|}{\multirow{13}{*}{}} & \multicolumn{1}{l}{Qwen2-72B-Instruct} & 57.0 & 82.9 & 67.6 & 37.7 & 70.9 & 49.2 & 34.0 & 55.2 & 42.1 & 32.3 & 53.1 & 40.2 \\
\multicolumn{1}{c|}{\multirow{13}{*}{}} & \multicolumn{1}{l}{Qwen2.5-7B-Instruct} & 40.6 & 33.2 & 36.5 & 30.8 & 45.1 & 36.6 & 26.5 & 33.9 & 29.7 & 26.2 & 28.6 & 27.4\\ 
\multicolumn{1}{c|}{\multirow{13}{*}{}} &\multicolumn{1}{l}{ Qwen2-7B-Instruct} & 40.6 & 4.7 & 8.4 & 30.5 & 13.8 & 19.0 & 22.4 & 10.9 & 14.7 & 20.0 & 8.7 & 12.1 \\
\multicolumn{1}{c|}{\multirow{13}{*}{}} &\multicolumn{1}{l}{ Llama-3.1-8B-Instruct} & 44.4 & 6.2 & 10.9 & 41.9 & 2.7 & 5.1 & 32.4 & 1.5 & 2.8 & 32.0 & 0.8 & 1.6 \\
\multicolumn{1}{c|}{\multirow{13}{*}{}} &\multicolumn{1}{l}{Meta-Llama-3-8B-Instruct} & 42.5 & 7.8 & 13.1 & 28.6 & 9.1 & 13.8 & 27.1 & 2.7 & 4.8 & 26.1 & 8.3 & 12.6 \\

\hline

   % \bottomrule          
\end{tabular}
\vspace{-5pt}
}

\end{table*}

\begin{table*}[h]
\centering    

\vspace{-10pt}
\caption{Breakdown of evaluation results of different variants of RetrievalPRM on ProcessBench.  We remove different components of RetrievalPRM to evaluate the contribution of each part to the model. The best result is given in bold, and the second-best value is underlined.
}
\vspace{-5pt}

\label{tab:ablation performance addition}
\resizebox{1.0\textwidth}{!}{
\renewcommand\arraystretch{1.1}
\begin{tabular}{cccccccccccccccc}
% \toprule
\hline

\multicolumn{2}{c}{Retrieval Components} & \multicolumn{3}{c}{GSM8k} &\multicolumn{3}{c}{MATH} &\multicolumn{3}{c}{OlympiadBench}&\multicolumn{3}{c}{OmniMATH}&\multirow{2}{*}{Avg.F1}\\ 

 \cmidrule(r){3-5} \cmidrule(r){6-8}  \cmidrule(r){9-11}  \cmidrule(r){12-14}
Question-level &Step-level & error  & correct & F1 & error  & correct & F1 & error  & correct & F1 & error  & correct & F1 & \\ 
   \hline

\checkmark &\checkmark& 64.7 &88.1&\underline{74.6} &67.2 & 75.6& \underline{71.1 }& 56.0 & 65.2&\textbf{ 60.2}&52.8 & 62.65& \textbf{57.33}&\textbf{65.8}\\
\checkmark&$\times$ &61.8& 94.8 &\textbf{74.9} &62.1&83.3 &\textbf{71.2}& 48.7&77.3&\underline{59.8} & 43.2 &73.4 &54.4&\underline{65.0}\\
$\times$& \checkmark& 51.7 &97.4 &67.5 &57.2 &87.4 & 69.2&46.0 & 82.0& 58.9& 43.9&78.4 & \underline{56.3} &63.0\\
$\times$&$\times$&50.7 & 92.7&65.6 & 57.9 & 81.0 & 67.5&46.9&68.7 & 55.8 &39.7& 71.0&50.9 &59.9\\

\hline   
\end{tabular}
\vspace{-5pt}
}

\end{table*}

% \section{Example Appendix}
\label{sec:appendix}

% This is an appendix.

\end{document}